\documentclass[conference]{IEEEtran}

\IEEEoverridecommandlockouts

\usepackage{cite}
\usepackage{amsmath,amssymb,amsfonts}
\usepackage{graphicx}
\usepackage{textcomp}
\usepackage{xcolor}
\usepackage{algorithm}
\usepackage{algpseudocode}
\usepackage{tikz}
\usetikzlibrary{shapes, arrows.meta, positioning}

\usepackage{xcolor}
\colorlet{black}{black}
\colorlet{myred}{red!80!black}
\colorlet{myblue}{blue!80!black}
\colorlet{mygreen}{green!60!black}
\colorlet{myorange}{orange!70!red!60!black}
\colorlet{mydarkred}{red!30!black}
\colorlet{mydarkblue}{blue!40!black}
\colorlet{mydarkgreen}{green!30!black}

\tikzset{
  >=latex,
  node/.style={thick,circle,draw=black,minimum size=22,inner sep=0.5,outer sep=0.6, text=black},
  node in/.style={node, fill=black!10, text=black, draw=black},
  node hidden/.style={node, fill=black!10, text=black, draw=black},
  node convol/.style={node, fill=black!10, text=black, draw=black},
  node out/.style={node, fill=black!10, text=black, draw=black},
  connect/.style={thick, black},
  connect arrow/.style={-{Latex[length=4,width=3.5]},thick,black,shorten <=0.5,shorten >=1},
  node 1/.style={node in},
  node 2/.style={node hidden},
  node 3/.style={node out},
  every picture/.style={color=black},
  every node/.style={text=black},
  every path/.style={color=black}
}

\IEEEoverridecommandlockouts

\usepackage{amsmath,amssymb,amsfonts}
\usepackage{graphicx}
\usepackage{xcolor}
\usepackage{tikz}
\usepackage{listofitems}
\usetikzlibrary{shapes, arrows.meta, positioning}

\colorlet{black}{black}
\colorlet{myred}{red!80!black}
\colorlet{myblue}{blue!80!black}
\colorlet{mygreen}{green!60!black}
\colorlet{myorange}{orange!70!red!60!black}
\colorlet{mydarkred}{red!30!black}
\colorlet{mydarkblue}{blue!40!black}
\colorlet{mydarkgreen}{green!30!black}

\tikzset{
  >=latex,
  node/.style={thick,circle,draw=black,minimum size=22,inner sep=0.5,outer sep=0.6, text=black},
  node in/.style={node, fill=black!10, text=black, draw=black},
  node hidden/.style={node, fill=black!10, text=black, draw=black},
  node convol/.style={node, fill=black!10, text=black, draw=black},
  node out/.style={node, fill=black!10, text=black, draw=black},
  connect/.style={thick, black},
  connect arrow/.style={-{Latex[length=4,width=3.5]},thick,black,shorten <=0.5,shorten >=1},
  node 1/.style={node in},
  node 2/.style={node hidden},
  node 3/.style={node out},
  every picture/.style={color=black},
  every node/.style={text=black},
  every path/.style={color=black}
}
\begin{document}

\title{Boosting	Hierarchical	Reinforcement	Learning	with	Meta-Learning	for Complex	Task Adaptation}

\author{\IEEEauthorblockN{Arash Khajooeinejad}
\IEEEauthorblockA{\textit{Electrical Engineering Department} \\
\textit{Iran University of Science and Technology}\\
Tehran, Iran \\
arash.khajooei@gmail.com}
\and
\IEEEauthorblockN{Fatemeh Sadat Masoumi†*}
\IEEEauthorblockA{\textit{Department of Computer Science
} \\
\textit{University of Texas at San Antonio}\\
San Antonio, Texas, USA \\
fatemeh.masoumi@my.utsa.edu}
\and
\IEEEauthorblockN{Masoumeh Chapariniya}
\IEEEauthorblockA{\textit{Institute of Computational Linguistics} \\
\textit{University of Zurich}\\
Zurich, Switzerland \\
masoumeh.chapariniya@uzh.ch}}

\maketitle

\begin{abstract}
Hierarchical	Reinforcement	Learning	(HRL)	is	well-suited	for	solving	complex	
tasks	by	breaking	them	down	into	structured	policies.	However,	HRL	agents	often	struggle	with	efficient	exploration	and	quick	adaptation.	To	overcome	these	limitations,	we	propose integrating	meta-learning	into	HRL	to	enable	agents	to	learn	and	adapt	hierarchical	policies	more	effectively.	Our	method	leverages	meta-learning	to	facilitate	rapid	task	adaptation	using	prior	experience,	while	intrinsic	motivation	mechanisms	drive	efficient	exploration	by	rewarding	the	discovery	of	novel	states.	
Specifically,	our	agent	employs	a	high-level	policy	to	choose	among	multiple	low-level	policies,	which	operate	within	custom-designed	grid	environments.	By	incorporating	gradient-based	meta-learning	with	differentiable	inner-loop updates,	we	optimize	performance	across	a	curriculum	of	progressively	challenging	tasks.	Experimental	results	highlight	that	our	metalearning-enhanced	hierarchicalagent	significantly	outperforms	standard	HRL	approaches	lacking	meta-learning	and	intrinsic	motivation.	
The	agent	demonstrates	faster	learning,	greater	cumulative	rewards,	and	higher	success	rates	in	complex	grid-based	scenarios.	These	Findings	underscore	the	
effectiveness	of	combining	meta-learning,	curriculum learning,	and	intrinsic	
motivation	to	enhance	the	capability	of	HRL	agents	in	tackling	complex	tasks.
\end{abstract}


\section{Introduction}
Reinforcement Learning (RL) has achieved success in domains such as gaming, robotics, and autonomous navigation by enabling agents to learn optimal policies through environment interactions \cite{Sutton2018}. However, traditional RL struggles with high-dimensional tasks, long-term dependencies, and sparse rewards due to the curse of dimensionality, making efficient exploration and learning challenging \cite{Lillicrap2016}.

Hierarchical Reinforcement Learning (HRL) addresses these challenges by decomposing tasks into subtasks, enabling agents to operate at different temporal levels \cite{Sutton1999, Barto2003}. Frameworks like the Options Framework \cite{Sutton1999} and Feudal Reinforcement Learning \cite{Vezhnevets2017} facilitate reusable sub-policy learning, improving exploration efficiency and mitigating dimensionality issues. Despite advancements, HRL agents face difficulties in exploring state spaces and adapting to novel tasks, particularly with sparse or deceptive rewards.

\textit{Meta-learning}, or “learning to learn,” enhances adaptability by enabling rapid policy adjustments based on prior experience \cite{Finn2017, Wang2016}. Integrating meta-learning into HRL allows agents to optimize both high- and low-level policies, improving learning efficiency across diverse tasks \cite{Frans2018, Sohn2018}. Intrinsic motivation mechanisms, such as curiosity-driven \cite{Pathak2017} and count-based exploration \cite{Bellemare2016}, further address exploration challenges by providing rewards for novel states or prediction errors. Curriculum learning complements this by sequencing tasks with increasing difficulty, helping agents build foundational skills before tackling complex problems \cite{Bengio2009, Narvekar2020 }.
We propose a framework integrating meta-learning into HRL, augmented with intrinsic motivation and guided by curriculum learning. Our agent employs a high-level policy to select among low-level options in custom grid environments of varying complexities. Meta-learning optimizes the learning process using gradient-based updates \cite{Finn2017} while intrinsic motivation encourages effective exploration, preventing convergence to suboptimal policies.
Experimental results show our framework outperforms traditional HRL agents, achieving faster learning, greater rewards, and higher success rates in complex environments. This highlights the potential of combining meta-learning, intrinsic motivation, and curriculum learning for tackling advanced RL tasks.
This is how the rest of the paper is organized: In Section II, relevant literature is reviewed; in Section III, the technique is explained; in Section IV, experiments and results are presented; and in Section V, new study directions are suggested.
\section{Related Work}
Hierarchical Reinforcement Learning (HRL) addresses the challenges of scaling reinforcement learning to complex tasks. Sutton \textit{et al.}~\cite{Sutton1999} introduced the Options Framework, formalizing temporally extended actions, while Bacon \textit{et al.}~\cite{Bacon2017}  proposed the Option-Critic Architecture for end-to-end learning of internal policies and termination conditions, enabling effective option discovery without predefined subgoals.
Meta-learning, or "learning to learn," enhances adaptability and sample efficiency across tasks. Finn \textit{et al.}~\cite{Finn2017} introduced Model-Agnostic Meta-Learning (MAML), enabling quick adaptation to new tasks in both supervised and reinforcement learning domains. Building on this, Frans \textit{et al.}~\cite{Frans2018} proposed Meta Learning Shared Hierarchies (MLSH), which meta-learns policy hierarchies for rapid adaptation in multi-task settings. Similarly, Houthooft \textit{et al.}~\cite{Houthooft2018} combined meta-learning with evolutionary strategies to develop adaptable policies using Evolved Policy Gradients.
Recent works have explored integrating meta-learning into HRL to tackle complex tasks. RL$^3$~\cite{Bhatia2023} combines traditional RL and meta-RL, excelling in long-horizon and out-of-distribution tasks, though it requires careful tuning. Meta Reinforcement Learning with Successor Feature-Based Context  ~\cite{Han2022} improves multi-task learning and rapid adaptation using context variables and successor features but faces scalability challenges due to reward decomposition complexity. Jiang \textit{et al.}~\cite{Jiang2021} introduced a context-based framework dividing learning into task inference and execution, enhancing exploration and sample efficiency but struggling with out-of-distribution tasks. Hierarchical Planning Through Goal-Conditioned Offline RL ~\cite{Li2022} and Variational Skill Embeddings for Meta-RL ~\cite{Chien2023} address long-horizon tasks and skill generalization, respectively, but both face limitations in real-time adaptability.
Intrinsic motivation addresses exploration challenges, with curiosity-driven methods ~\cite{Pathak2017} and pseudo-count-based methods ~\cite{Bellemare2016} guiding agents toward novel states. Curriculum learning, as surveyed by Bengio \textit{et al.}~\cite{Bengio2009} and Narvekar \textit{et al.}~\cite{Narvekar2020}, improves learning by structuring tasks progressively. Integration of HRL with intrinsic motivation, such as in Hierarchical DQN ~\cite{Kulkarni2016} and FeUdal Networks (FuN) ~\cite{Vezhnevets2017}, further enhances exploration and efficiency in complex environments.
Despite significant progress in HRL, meta-RL, and intrinsic motivation, integrating these methods holistically to enhance learning remains a challenge. Our approach addresses this by combining meta-learning, intrinsic motivation, and curriculum learning within an HRL framework. Unlike prior works like ~\cite{Frans2018} and RL$^3$~\cite{Bhatia2023}, which focus on specific aspects, our method applies meta-learning to both high- and low-level policies. This enables rapid adaptation to tasks of varying complexities, improving exploration, adaptability, and learning efficiency in both short- and long-term scenarios.
We incorporate intrinsic motivation—extending beyond curiosity-driven and count-based methods ~\cite{Pathak2017, Bellemare2016} to encourage exploration of novel states and overcome local minima. Combined with curriculum learning inspired by Bengio \textit{et al.}~\cite{Bengio2009} and Florensa \textit{et al.}~\cite{Florensa2017}, tasks are structured progressively to build foundational skills incrementally. This integrated approach enables efficient exploration, rapid adaptation, and superior performance in complex environments.
\section{Methodology}
Our proposed methodology integrates multiple advanced learning techniques, including Hierarchical Reinforcement Learning (HRL), meta-learning, intrinsic motivation, and curriculum learning. Each component is designed to address specific challenges such as scalability, rapid adaptation, efficient exploration, and learning complex tasks with sparse rewards. This section provides a detailed explanation of each element in the framework, how they interact, and their theoretical foundations.

\subsection{Overall Framework}
Our framework employs Hierarchical Reinforcement Learning (HRL), decomposing the policy into high-level and low-level components. The high-level policy selects abstract actions, or \textit{options}, while low-level policies execute primitive actions. This structure introduces temporal abstraction, enabling decisions over multiple time steps instead of at each step.
Options in HRL, as introduced by Sutton \textit{et al.} \cite{Sutton1999}, consist of an \textit{initiation set} \( \mathcal{I}_\omega \), defining states where an option can begin; an \textit{intra-option policy} \( \pi_\omega \), mapping states to actions; and a \textit{termination function} \( \beta_\omega \), which determines when the option ends and control returns to the high-level policy. By executing intra-option policies until termination, the agent focuses on sub-goals within a larger task.
The high-level policy \( \pi_h(\omega \mid s) \) selects options based on the current state, determining \textit{which} option to execute, while the low-level policy \( \pi_\omega(a \mid s) \) dictates \textit{how} to act during the option’s execution. Temporal abstraction reduces task complexity by enabling multi-step decision-making, enhancing the agent’s efficiency in handling complex tasks.
\begin{figure}[htbp]
    \centering
    \begin{tikzpicture}[scale=0.8, transform shape,
        node distance=0.8cm and 1cm, 
        every node/.style={align=center, font=\sffamily, text=black},
        every path/.style={draw=black, opacity=1},
        box/.style={draw, minimum width=2cm, minimum height=0.8cm, rounded corners, text centered, font=\scriptsize},
        process/.style={draw, rectangle, minimum width=2cm, minimum height=0.8cm, rounded corners, fill=blue!20, text centered, font=\scriptsize},
        highlevel/.style={draw, rectangle, rounded corners, fill=green!30, minimum height=0.8cm, minimum width=2.5cm, text centered, font=\scriptsize, text=black},
        lowlevel/.style={draw, rectangle, rounded corners, fill=orange!30, minimum height=0.8cm, minimum width=2.5cm, text centered, font=\scriptsize, text=black},
        decision/.style={diamond, draw, aspect=2, text centered, inner sep=0pt, minimum width=1.5cm, font=\scriptsize, text=black},
        arrow/.style={-Latex, thick, draw=black, opacity=1},
        dashedarrow/.style={-Latex, thick, draw=black, dashed, opacity=1}
    ]

    \node (high_policy) [highlevel] {High-Level Policy \\ \( \pi_h(s_t) \)};

    \node (low_policy_1) [lowlevel, below left=of high_policy] {Low-Level Policy \\ \( \pi_{\omega_1}(s_t) \)};
    \node (low_policy_2) [lowlevel, below=of low_policy_1] {Low-Level Policy \\ \( \pi_{\omega_2}(s_t) \)};
    \node (low_policy_n) [lowlevel, below=of low_policy_2] {Low-Level Policy \\ \( \pi_{\omega_n}(s_t) \)};

    \node (env) [process, right=of low_policy_2, xshift=0.5cm] {Environment}; 

    \node (decision) [decision, below=of high_policy, yshift=-0.2cm] {Select Option \\ \( \omega_t \)};

    \draw [arrow] (high_policy) -- (decision);

    \draw [arrow] (decision) -- (low_policy_1);
    \draw [arrow] (decision) -- (low_policy_2);
    \draw [arrow, dotted, draw=black] (decision) -- (low_policy_n);

    \draw [arrow] (low_policy_1) -- (env);
    \draw [arrow] (low_policy_2) -- (env);
    \draw [arrow, dotted, draw=black] (low_policy_n) -- (env);

    \draw [dashedarrow] (env.east) .. controls +(1.5,0) and +(1.5,0) .. node [above, text=black] {State \( s_{t+1} \)} (high_policy.east);

    \end{tikzpicture}
    \caption{Architecture for Hierarchical Reinforcement Learning (HRL). After the high-level policy makes a choice, a low-level policy is triggered to engage with the environment. Based on input from the surroundings, the agent continuously modifies its state. The hierarchical flow of feedback and decision-making between high-level and low-level policies is shown by the dashed lines.}
    \label{fig:HRL_architecture}
\end{figure}

\begin{figure}[h!]
    \centering
    \begin{tikzpicture}[scale=0.8, transform shape,
    node distance=0.5cm, 
    every node/.style={align=center, font=\scriptsize}, 
    box/.style={rectangle, draw, minimum width=2.5cm, minimum height=0.8cm, rounded corners},
    process/.style={draw, rectangle, minimum width=2.5cm, minimum height=0.8cm, rounded corners, fill=blue!20},
    state/.style={draw, rectangle, rounded corners, fill=green!20, minimum height=0.8cm, minimum width=2.5cm},
    reward/.style={draw, rectangle, rounded corners, fill=yellow!20, minimum height=0.8cm, minimum width=2.5cm},
    env/.style={draw, rectangle, rounded corners, fill=purple!20, minimum height=0.8cm, minimum width=2.5cm},
    arrow/.style={-Latex, thick}
]

    \node (agent) [box] {Agent};
    
    \node (env) [env, right=of agent, xshift=0.5cm] {Environment};
    
    \node (state_visitation) [process, below=of env, yshift=-0.2cm] {State Visitation Count \\ \( N(s_t) \)};
    
    \node (intrinsic_reward) [reward, below=of state_visitation, yshift=-0.2cm] {Intrinsic Reward \\ \( r^{\text{int}}_t = \eta \cdot \frac{1}{\sqrt{N(s_t) + \epsilon}} \)};
    
    \node (total_reward) [reward, left=of intrinsic_reward, xshift=-0.5cm] {Total Reward \\ \( r^{\text{total}}_t = r^{\text{ext}}_t + r^{\text{int}}_t \)};
    
    \node (next_state) [state, above=of agent, yshift=0.2cm] {Next State \\ \( s_{t+1} \)};
    
    \draw [arrow] (agent) -- (env) node[midway, above] {Action \( a_t \)};
    \draw [arrow] (env) -- (state_visitation) node[midway, right] {State \( s_t \)};
    \draw [arrow] (state_visitation) -- (intrinsic_reward);
    \draw [arrow] (intrinsic_reward) -- (total_reward);
    \draw [arrow] (total_reward) -- (agent);
    \draw [arrow, dashed] (env) -- (next_state) node[midway, right] {New State \( s_{t+1} \)};
    \draw [arrow, dashed] (next_state) -- (agent);

\end{tikzpicture}
    \caption{Intrinsic Motivation and Exploration Path: This flowchart illustrates how the agent interacts with the environment, calculates intrinsic rewards based on state visitation counts, and uses a combination of intrinsic and extrinsic rewards to guide exploration. The feedback loop ensures that new states are visited and counted, promoting efficient exploration.}
    \label{fig:intrinsic_motivation}
\end{figure}

\paragraph{Workflow of the Agent}
The agent's decision-making follows a hierarchical structure Algorithm ~\ref{alg:decision_process}. The high-level policy selects options based on the current state, while the low-level policy executes actions within the option until reaching a goal state or meeting termination conditions. This framework enables long-term goal targeting by the high-level policy and short-term execution by the low-level policies, with feedback loops guiding future option selections.
\begin{algorithm}[h]
\scriptsize
\caption{Hierarchical Decision-Making Process}\label{alg:decision_process}
\begin{algorithmic}[1]
\Require Initial environment state \( s_0 \)
\Ensure Episode completion
\State Initialize time step \( t \gets 0 \)
\While{episode not terminated}
    \If{option terminated or at initial step}
        \State Select option \( \omega_t \sim \pi_h(\cdot \mid s_t) \) \Comment{High-Level Decision}
    \EndIf
    \State Select action \( a_t \sim \pi_\omega(\cdot \mid s_t) \) \Comment{Low-Level Action}
    \State Execute action \( a_t \), observe reward \( r_t \), and next state \( s_{t+1} \)
    \State Accumulate reward \( R_t \gets R_t + r_t \)
    \State \textbf{Option Termination Check}: With probability \( \beta_\omega(s_{t+1}) \), set \textit{option terminated} \( \gets \) True
    \State Update state \( s_t \gets s_{t+1} \)
    \State \( t \gets t + 1 \)
\EndWhile
\end{algorithmic}
\end{algorithm}

This hierarchical structure enables the agent to handle temporally extended actions, with the high-level policy targeting long-term goals and low-level policies managing short-term execution. The dashed arrow from the environment to the high-level policy represents the feedback loop, where updated states guide future option selections.
\subsection{Meta-Learning Integration}
The key challenge in reinforcement learning is the need to rapidly adapt to new tasks or environments. Meta-learning, also known as ``learning to learn,'' provides a solution by optimizing the agent’s ability to adapt. In our framework, meta-learning is applied to both the high-level and low-level policies.
\paragraph{Meta-Learning Framework}
In meta-learning, the goal is to learn parameters \( \theta \) that allow rapid adaptation to new tasks with only a few updates. We employ a gradient-based meta-learning approach inspired by Model-Agnostic Meta-Learning (MAML) \cite{Finn2017}. This framework involves an \textit{inner loop}, where task-specific learning occurs, and an \textit{outer loop}, where meta-parameters are updated across tasks.
\begin{figure}[h!]
    \centering
    \begin{tikzpicture}[node distance=1.0 cm, every node/.style={align=center, font=\tiny}]
\node (meta_init) [rectangle, draw, rounded corners, text centered, minimum width=2.5cm, minimum height=0.6cm, font=\tiny] {Meta-parameters \\ Initialization};

\node (sample_task) [rectangle, draw, below of=meta_init, rounded corners, text centered, minimum width=2cm, minimum height=0.6cm, font=\tiny] {Sample Task \\ from Distribution \\ \( \mathcal{T} \)};

\node (inner_update) [rectangle, draw, below of=sample_task, rounded corners, text centered, minimum width=2cm, minimum height=0.6cm, font=\tiny] {Inner-loop Update \\ Task \\ \( \mathcal{T}_i \)};

\node (adapt_params) [rectangle, draw, below of=inner_update, rounded corners, text centered, minimum width=2cm, minimum height=0.6cm, font=\tiny] {Compute Adapted \\ Parameters \\ \( \theta'_{\mathcal{T}_i} \)};

\node (meta_loss) [rectangle, draw, below of=adapt_params, rounded corners, text centered, minimum width=2cm, minimum height=0.6cm, font=\tiny] {Compute Meta-loss \\ \( \mathcal{L}_{\text{meta}} \)};

\node (meta_update) [rectangle, draw, below of=meta_loss, rounded corners, text centered, minimum width=2cm, minimum height=0.6cm, font=\tiny] {Update Meta-parameters \\ \( \theta \gets \theta - \beta \nabla \mathcal{L}_{\text{meta}} \)};

\node (new_task) [draw=none, below of=meta_update, node distance=0.5cm, text centered, font=\tiny] {Repeat for Next Task};

        \draw[->] (meta_init) -- (sample_task);
        \draw[->] (sample_task) -- (inner_update);
        \draw[->] (inner_update) -- (adapt_params);
        \draw[->] (adapt_params) -- (meta_loss);
        \draw[->] (meta_loss) -- (meta_update);
        \draw[->] (meta_update) -- (new_task);
    
    \end{tikzpicture}
    \caption{Meta-Learning Process Flowchart. The outer loop initializes and updates meta-parameters across tasks, while the inner loop performs task-specific adaptations using gradient descent. The meta-loss is computed based on adapted parameters to optimize the meta-parameters.}
    \label{fig:meta_learning_flowchart}
\end{figure}

\paragraph{Meta-Parameters and Task Distribution}

In our hierarchical framework, the meta-parameters \( \theta \) include \( \theta_h \), the parameters of the high-level policy \( \pi_h \); \( \theta_\omega \), the parameters of the intra-option policy for each option \( \omega \); and \( \theta_{\beta_\omega} \), the parameters of the termination function \( \beta_\omega \).
The agent is trained on a distribution of tasks \( \mathcal{T} \), and the objective is to find meta-parameters that can be quickly adapted for each task.
\paragraph{Meta-Learning Objective}
The meta-learning objective is to minimize the expected loss over the task distribution \( \mathcal{T} \):
\begin{equation}
\min_{\theta} \mathbb{E}_{\mathcal{T}_i \sim \mathcal{T}} \left[ \mathcal{L}_{\mathcal{T}_i} \left( \theta'_{\mathcal{T}_i} \right) \right], \tag{2}
\end{equation}
where \( \theta'_{\mathcal{T}_i} \) are the adapted parameters for task \( \mathcal{T}_i \), obtained after performing \( K \) inner-loop updates using task-specific data. The inner-loop updates are performed using gradient descent:
\begin{equation}
\theta'_{\mathcal{T}_i} = \theta - \alpha \nabla_{\theta} \mathcal{L}_{\mathcal{T}_i}(\theta), \tag{3}
\end{equation}
where \( \alpha \) is the learning rate for inner-loop updates. The meta-parameters \( \theta \) are updated in the outer loop, based on the loss computed after adaptation to each task.
\paragraph{Meta-Training Algorithm}
The meta-training process integrates hierarchical reinforcement learning, meta-learning, intrinsic motivation, and curriculum learning. This ensures rapid adaptation to new tasks by updating both high-level and low-level policies, as detailed in ~\ref{alg:meta_training}.
\begin{algorithm}[h]
\scriptsize
\caption{Meta-Training Procedure}\label{alg:meta_training}
\begin{algorithmic}[1]
\Require \(\theta\), rates \(\alpha, \beta\), levels \(\{\ell\}\)
\Ensure Optimized \(\theta\)
\State \(\ell \gets 1\)
\For{each meta-iteration}
    \State Sample \(\{\mathcal{T}_i\}\) from level \(\ell\)
    \For{each \(\mathcal{T}_i\)}
        \State \(\theta'_{\mathcal{T}_i} \gets \theta\)
        \For{each inner step}
            \State Reset task environment
            \State Collect trajectories using \(\theta'_{\mathcal{T}_i}\)
            \State Update \(N(s)\), compute \(r^{\text{int}}_t\)
            \State Compute losses \(\mathcal{L}^{\text{high}}_{\mathcal{T}_i}, \mathcal{L}^{\text{low}}_{\mathcal{T}_i}, \mathcal{L}^{\beta}_{\mathcal{T}_i}\)
            \State \(\theta'_{\mathcal{T}_i} \gets \theta'_{\mathcal{T}_i} - \alpha \nabla_{\theta'_{\mathcal{T}_i}} (\mathcal{L}^{\text{high}}_{\mathcal{T}_i} + \sum_{\omega} (\mathcal{L}^{\text{low}}_{\mathcal{T}_i} + \mathcal{L}^{\beta}_{\mathcal{T}_i}))\)
        \EndFor
        \State Compute \(\mathcal{L}^{\text{meta}}_{\mathcal{T}_i}\) using \(\theta'_{\mathcal{T}_i}\)
    \EndFor
    \State \(\theta \gets \theta - \beta \nabla_{\theta} \sum_{\mathcal{T}_i} \mathcal{L}^{\text{meta}}_{\mathcal{T}_i}\)
    \If{performance meets threshold} \(\ell \gets \ell + 1\) \EndIf
\EndFor
\end{algorithmic}
\end{algorithm}
\subsection{Neural Network Architectures}
\begin{figure}[!t]
    \centering
    \includegraphics[width=\columnwidth]{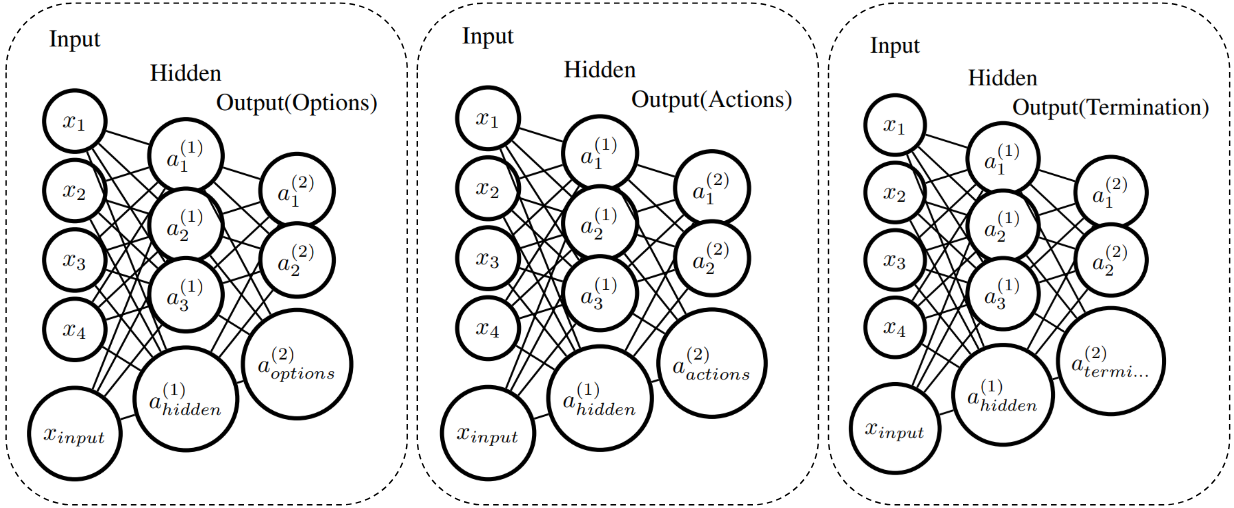}
    \caption{Neural network architectures for the three key components of the hierarchical reinforcement learning system: (1) High-Level Policy Network (Options), (2) Low-Level Policy Network (Actions), and (3) Termination Function Network. Each network consists of an input layer, hidden layers (64 and 32 neurons), and an output layer tailored to the respective tasks.}
    \label{fig:neural_network_architectures}
\end{figure}

Our framework employs three neural networks to support the high-level policy, low-level policy, and termination function, enabling decision-making at different abstraction levels and adaptability to complex environments (Figure \ref{fig:neural_network_architectures}, left).

The High-Level Policy Network selects options guiding actions over extended time horizons. It takes the current state as a one-hot encoded vector (e.g., 36 states for a 6×6 grid) and processes it through three fully connected layers: an input layer, two hidden layers (64 and 32 units, ReLU activations), and an output layer representing 5 possible options.
The Low-Level Policy Network determines actions within the chosen high-level option. Similar in structure to the high-level network, it processes the state as a one-hot encoded vector through two hidden layers (64 and 32 units, ReLU activations) and outputs action values for predefined primitive actions (up, down, left, right).
The Termination Function Network determines when to end a high-level option. It processes the current state as a one-hot encoded input through two hidden layers (64 and 32 units, ReLU activations) and outputs termination probabilities via a sigmoid activation, enabling smooth transitions between high-level strategies.

The three networks are optimized with Adam for meta-learning and SGD for inner-loop adaptation. Joint training of high-level policies, low-level policies, and the termination function maximizes rewards, enabling dynamic option switching, balancing exploration and exploitation, and tackling complex tasks with sparse rewards.
\subsection{Intrinsic Motivation Mechanism}
To enhance exploration in complex environments with sparse rewards, we use an intrinsic motivation mechanism based on state visitation counts. This provides rewards for visiting less frequently explored states, encouraging the agent to discover novel states efficiently.
\paragraph{Intrinsic Reward Formulation}
The intrinsic reward at time \( t \) is defined as:
\begin{equation}
r^{\text{int}}_t = \eta \cdot \frac{1}{\sqrt{N(s_t) + \epsilon}}, \tag{4}
\end{equation}
where \( \eta \) is a scaling factor that controls the magnitude of intrinsic rewards.
 \( N(s_t) \) is the number of times the agent has visited state \( s_t \).
 \( \epsilon \) is a small constant to prevent division by zero.
By combining the extrinsic reward from the environment with the intrinsic reward from exploration, the agent’s total reward becomes:
\begin{equation}
r^{\text{total}}_t = r^{\text{ext}}_t + r^{\text{int}}_t. \tag{5}
\end{equation}
This ensures that the agent balances between exploring new areas of the state space and exploiting known strategies to complete the task.

\paragraph{Exploration Path Visualization}
Figure~\ref{fig:intrinsic_motivation} illustrates how intrinsic motivation drives exploration, showing the agent interacting with the environment, receiving rewards from state visitation counts, and updating its path based on combined intrinsic and extrinsic rewards.
\subsection{Curriculum Learning Strategy}
We implement a curriculum learning strategy that gradually increases task difficulty as the agent improves. This structured progression helps the agent build foundational skills on simpler tasks before addressing more complex ones.
Each curriculum level is characterized by grid size, number of traps  (obstacles or traps in the environment), and task complexity  (difficulty based on path length, trap density, and goal distance).
\paragraph{Performance-Based Progression}
The agent advances to harder curriculum levels upon reaching performance thresholds, such as success rate or cumulative reward. This adaptive difficulty prevents early overwhelm, fostering gradual skill acquisition and efficient learning.
\subsection{Policy Optimization with Intrinsic Rewards}
The agent’s policies are optimized using Q-learning, where the Q-values are updated based on both extrinsic and intrinsic rewards. The target Q-value for the intra-option policy is adjusted to account for the total reward:
\begin{equation}
y^{\text{low}}_t = r^{\text{total}}_t + \gamma \max_{a'} Q^{\pi_\omega}(s_{t+1}, a'; \theta_\omega^{-}), \tag{6}
\end{equation}
where \( \gamma \) is the discount factor, and \( \theta_\omega^{-} \) represents the parameters of a target network used for stabilization during training. The intrinsic rewards encourage exploration, while the extrinsic rewards guide the agent toward task completion.
\section{Experimental Setup}
\label{sec:experimental_setup}
This section outlines the environments, baseline comparisons, hyperparameter optimization using Optuna, evaluation metrics, and computational resources. Experiments were conducted in two scenarios: a fixed complexity environment and a curriculum learning setup with gradually increasing complexity. 
Custom grid-based environments were used to simulate navigation tasks, varying in grid size and the number of traps that act as obstacles for the agent to navigate while reaching its goal.
\subsection{Experimental Scenarios}
\label{subsec:experimental_scenarios}
We evaluated the performance and adaptability of the proposed meta-learning integrated HRL framework under three scenarios. First, hyperparameter optimization and validation were performed using {Optuna} \cite{Akiba2019}, an automatic hyperparameter optimization library. Optuna conducted 50 trials with the {MedianPruner}, pruning unpromising trials early to improve efficiency. Each trial trained the agent with specific hyperparameters, evaluating performance based on the average reward over the last 10 meta-iterations. The most optimal trial yielded the hyperparameters shown in Table I. 
Second, the framework was trained and evaluated in a stable environment with constant complexity to assess its baseline performance. Lastly, we compared the agent's performance in static complex environments versus gradually increasing task complexity through curriculum learning, highlighting the framework's adaptability to dynamic challenges. Each scenario assessed different aspects of the framework, ranging from hyperparameter sensitivity to adaptability in varying environments.
\begin{table}[h!]
\centering
\caption{Optimal Hyperparameters Identified via Optuna Optimization}
\label{tab:optimal_hyperparameters}
\begin{tabular}{|c|c|}
\hline
\textbf{Parameter} & \textbf{Optimal Value} \\ \hline
Meta-Learning Rate (\(\beta\)) & \(8.24 \times 10^{-6}\) \\ \hline
Inner-Loop Learning Rate (\(\alpha\)) & 0.00317 \\ \hline
Number of Inner Steps & 5 \\ \hline
High-Level Exploration (\(\epsilon_{\text{high}}\)) & 0.1018 \\ \hline
Option Exploration (\(\epsilon_{\text{option}}\)) & 0.6199 \\ \hline
Intrinsic Reward Scale (\(\eta\)) & 0.1111 \\ \hline
\end{tabular}
\end{table}
\subsubsection{Fixed Complexity Scenario}
\label{subsubsec:fixed_complexity_scenario}
In this scenario, the agent is trained and evaluated in a stable environment with constant complexity, serving as a baseline to assess the effectiveness of the hierarchical and meta-learning components without the influence of increasing task difficulty.
\paragraph{Environment Configuration}
The grid size \(6 \times 6\) and number of traps are 3.
The environment's consistent difficulty ensures the agent’s learning process is unaffected by varying complexities, enabling isolated analysis of the hierarchical and meta-learning mechanisms.

The training parameters were set as follows: the meta-learning rate (\(\beta\)) was 0.0001, with an inner-loop learning rate (\(\alpha\)) of 0.003. The number of inner steps was set to 3, while the high-level exploration (\(\epsilon_{\text{high}}\)) and option exploration (\(\epsilon_{\text{option}}\)) were 0.3 and 0.5, respectively. The intrinsic reward scale (\(\eta\)) was fixed at 0.1. Additionally, the training involved 500 meta-iterations, with 50 inner steps per task.
These training hyperparameters configuration allows us to evaluate the agent's ability to learn and perform effectively without the added complexity of changing tasks.
In the fixed complexity scenario, the agent’s performance is assessed using metrics such as Meta-Loss, Average Reward, Success Rate, Exploration Efficiency, and Cumulative Rewards. Over 500 meta-iterations, meta-loss steadily decreases, converging around 30, while the average reward stabilizes at -5 after initial fluctuations. The success rate improves within 200 iterations but shows oscillations, reflecting a balance between exploration and exploitation. Overall, the scenario highlights gradual policy improvement, stabilized trends, and opportunities for optimizing consistent goal achievement.
\begin{figure}[!t]
    \centering
    \includegraphics[width=0.48\textwidth]{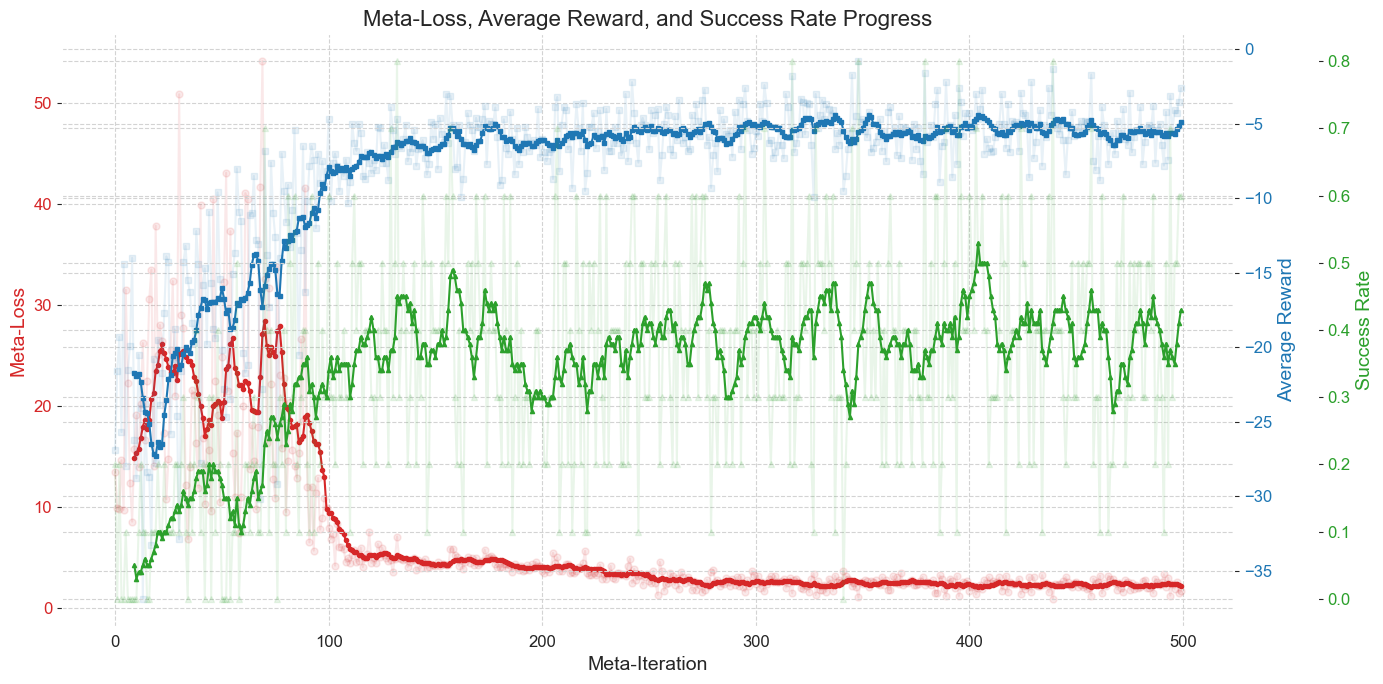}
    \caption{The graph shows the progression of Meta-Loss, Average Reward, and Success Rate over 500 meta-iterations in the fixed complexity scenario. The red line represents Meta-Loss, the blue line indicates Average Reward, and the green line shows Success Rate.}
    \label{fig:fixed_complexity_graph}
\end{figure}
\subsubsection{Gradual Complexity Scenario }
\label{subsubsec:gradual_complexity_results}
\begin{figure}[!t]
\centering
\includegraphics[width=0.48\textwidth]{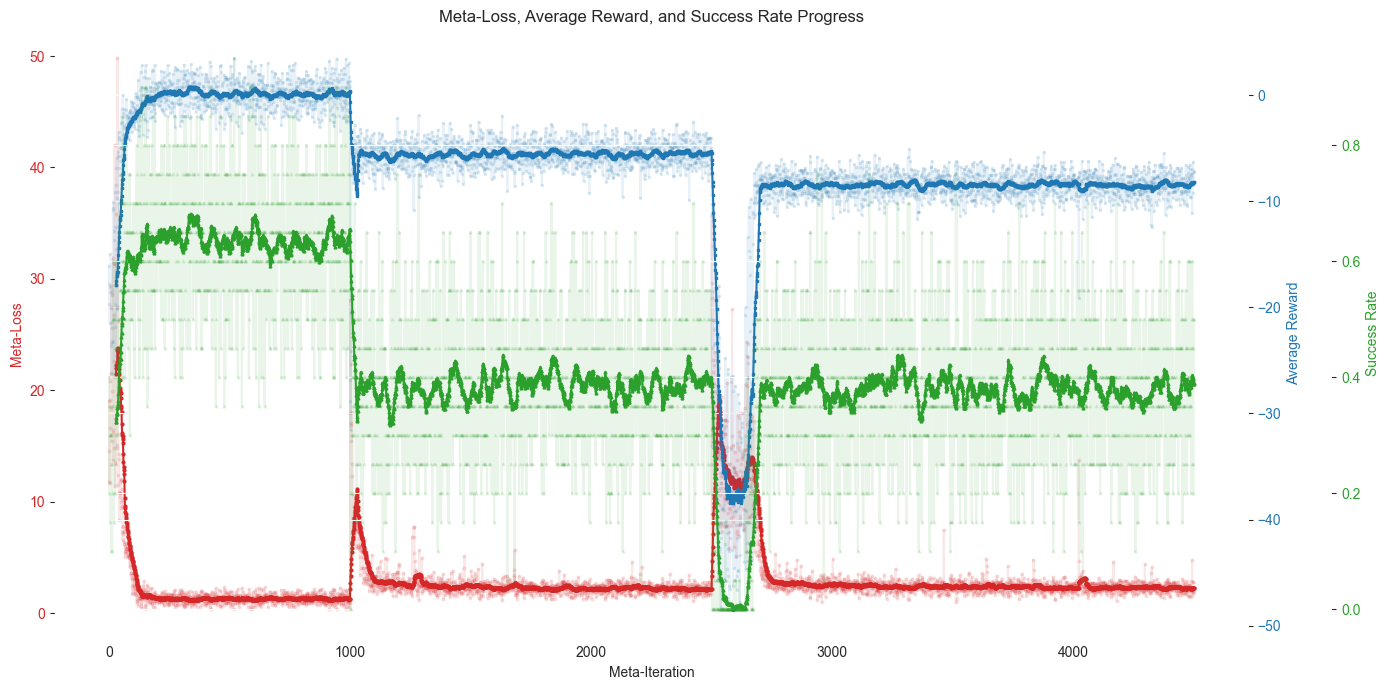}
\caption{The graph illustrates the progression of Meta-Loss, Average Reward, and Success Rate over 4000 meta-iterations in the gradual complexity scenario. The red line represents Meta-Loss, the blue line indicates Average Reward, and the green line shows Success Rate. }
\label{fig:gradual_complexity_graph}
\end{figure}

The gradual complexity scenario evaluates the agent's adaptability by progressively increasing task difficulty over 4000 meta-iterations (Figure~\ref{fig:gradual_complexity_graph}). Meta-loss fluctuates sharply during transitions but trends downward overall, indicating gradual refinement. Average rewards drop with increasing complexity but recover over time, while success rates stabilize at 50-60\% in simpler phases and decline during transitions, reflecting the balance between exploration and exploitation.
\paragraph{Comparison with Fixed Complexity Scenario}
Compared to the fixed complexity scenario, which achieves faster policy stabilization and a steady success rate (40--45\%), the gradual complexity scenario highlights the challenges of adapting to dynamic environments. While fixed complexity fosters stability, the gradual scenario tests adaptability by requiring the agent to relearn and adjust to harder tasks. These results emphasize the importance of dynamic policy mechanisms for handling increasing task complexity effectively.
\section{Conclusion}
We presented an enhanced hierarchical reinforcement learning framework integrating meta-learning, intrinsic motivation, and curriculum learning to improve adaptability, exploration, and performance in complex tasks. Experimental results showed faster convergence, higher success rates, and better adaptability compared to traditional methods, demonstrating the effectiveness of this approach for tackling dynamic and challenging environments.

\end{document}